\documentclass[sigconf,nonacm]{acmart}
\usepackage{graphicx}
\usepackage{subcaption}
\usepackage{xcolor}
\usepackage{lipsum}

\AtBeginDocument{%
  }

\begin{document}

\title[Neuro-Symbolic Traders]{Neuro-Symbolic Traders: Assessing the \\Wisdom of AI Crowds in Markets}

\author{Namid R. Stillman}
\email{namid@simudyne.com}
\affiliation{%
  \institution{Simudyne Limited}
  \country{United Kingdom}
}

\author{Rory Baggott}
\email{rory@simudyne.com}
\affiliation{%
  \institution{Simudyne Limited}
  \country{United Kingdom}
}


\begin{abstract}

Deep generative models are becoming increasingly used as tools for financial analysis. However, it is unclear how these models will influence financial markets, especially when they infer financial value in a semi-autonomous way. In this work, we explore the interplay between deep generative models and market dynamics. We develop a form of virtual traders that use deep generative models to make buy/sell decisions, which we term neuro-symbolic traders, and expose them to a virtual market. Under our framework, neuro-symbolic traders are agents that use vision-language models to discover a model of the fundamental value of an asset. Agents develop this model as a stochastic differential equation, calibrated to market data using gradient descent. We test our neuro-symbolic traders on both synthetic data and real financial time series, including an equity stock, commodity, and a foreign exchange pair. We then expose several groups of neuro-symbolic traders to a virtual market environment. This market environment allows for feedback between the traders belief of the underlying value to the observed price dynamics. We find that this leads to price suppression compared to the historical data, highlighting a future risk to market stability. Our work is a first step towards quantifying the effect of deep generative agents on markets dynamics and sets out some of the potential risks and benefits of this approach in the future. 
\end{abstract}

\maketitle

\section{Introduction}\label{sec:intro}

Isaac Newton is thought to have once remarked, in relation to his exposure to the  South Sea Bubble crisis, that he ``can calculate the motion of heavenly bodies, but not the madness of people" \cite{odlyzko2018isaac}. This comment reflects the observation that collective intelligence, where the opinions or beliefs of individuals are aggregated, can lead to either robust reasoning (the `wisdom of the crowds') or group-think and herding which, in turn, can increase the groups exposure to unexpected events \cite{kirman2010complex}. 

The benefits and risks of collective intelligence are evidenced by price fluctuations in financial markets. On the one hand, markets are thought to act as an effective information aggregation mechanism in determining the monetary value of an asset. However, large swings and boom-bust cycles are compounded by herding amongst traders or contagion across the collective financial network, which is directly related to the underlying aggregation process \cite{cipriani2008herd}.

Recent methods in machine learning and artificial intelligence (AI) have dramatically changed both models and mechanism of price discovery. This is due to increasingly large volumes of data that need to be processed quickly, the rise of automated trading systems, and the recent dominance of deep learning methods which use over-parameterised artificial neural networks (NNs) to identify signals without the reliance on traditional heuristics or domain-based assumptions. Given this, markets may be moving towards a more automated and statistical framework for price discovery \cite{kumiega2012automated}.

One of the most promising sub-fields of deep learning research is deep generative modelling. These methods use NNs to approximate the underlying data generation process and include methods such as normalising flows, diffusion models, and language and multi-modal models \cite{Tomczak2024DeepGenerative}. This latter class of language and multi-modal models have shown particular strength in assisting with technical analysis, especially within agentic workflows \cite{guo2024large}. 

Agentic workflows are those that design systems where language models act as semi-autonomous agents. In this work, we develop agents that use ``Box's loop" (see, for example, \cite{box1962useful, blei2014build}) to develop models of the fundamental value of a financial asset. We refer to these agents as \textit{neuro-symbolic traders} and expose them to a virtual market. The market environment allows feedback between the agents belief and the market price of the asset. This gives us a principled way to investigate how markets with artificial agent behave and is an preliminary step in the exploration of AI crowding in markets. We note that this work does not aim to develop either a robust prediction engine for price dynamics or a comprehensive neuro-symbolic framework containing the complete set of tools available in quantitative finance. Instead, we design this as as a toy environment to explore the effect of AI agents on price paths. 

Our work is set out as follows; we first review other relevant works. We then set out the methods used in this work, including the baseline models that we use for predicting the fundamental value of an asset, how these models are adapted and improved upon using vision-language models, and the structure of our market simulator which represents a simplified toy model of a market. We then give details on the different experiments we have conducted to test our methods, before describing our results and making concluding comments. Finally, we give a brief summary of what we believe to be the significance of our work. 

\section{Relevant Work}

Generative models provide a principled way to reproduce the data-generating process and form a critical part of the scientific process \cite{sankaran2023generative}. Broadly, generative models can be used to test hypotheses and the limits of understanding. Simulations are interpretable generative models based on a set of defined assumptions, while other models use artificial neural networks to increase performance at the expense of interpretability. In this work, we use deep generative models (vision-language models) to develop and discover traditional mathematical models (stochastic differential equations). This approach is inspired by the work set-out in \cite{li2024automated}, where language models act within a framework inspired by Box's loop \cite{box1962useful, blei2014build}. 

The use of generative models as tools for understanding price dynamics has a long history. Modelling price paths using stochastic differential equations may be considered an early example of a generative model \cite{meyer2009stochastic}. More recently, market simulators and deep generative models have been developed that outputs synthetic market data with statistical features matching historical observations \cite{jacobs2004financial, raberto2001agent}. These simulators have been used for risk management and for testing trading strategies. See, for example, \cite{axtell2022agent} for a recent review. However, at the time of writing, we are unaware of examples of market simulators which include traders that have their own generative models. 



The combination of different types of generative models has been explored in multiple scientific disciplines. These hybrid or \textit{neuro-symbolic} methods combine traditional symbolic generative methods with deep learning and seek to maintain the interpretability of symbolic models but augment unknown components with `black box' models defined by artificial neural networks (NNs) \cite{yin2021augmenting, cohen2023black, sheth2023neurosymbolic}. Language models represent a wholly new class  of neuro-symbolic methods, where the language models can be used to perform program search either by enumerating possible valid programs (brute force or system one thinking), or through a form of iterative discovery (system two thinking) \cite{nye2021improving,chollet2019measure}. Language models are particularly interesting as a neuro-symbolic method given their large training corpus. The generative output of these models can be considered as a form of interpolation across the various data points within the training set, analogous to collective reasoning \cite{chuang2024wisdom}. 

Hence, there is a direct link between language models and financial markets. They both can be described as performing collective averaging, where price signals in markets represent the aggregation of individual trader beliefs on the value of an asset \cite{kameda2022information, varian1989differences}. However, we are unaware on any work that explores the relationship between these aggregation methods. Given this, there is an open question as to what happens when neuro-symbolic methods are exposed to a market environment. In this work we set out to provide preliminary evidence on this, highlighting both the benefits and risks of markets driven by neuro-symbolic agents. Our work represents an early example of how neuro-symbolic methods, combining language models with traditional quantitative finance methods, can affect the market dynamics. However, we expect there is much left to explore within this space. 

\section{Methods}

\begin{figure}[!t]
\begin{center}
\includegraphics[width=8cm]{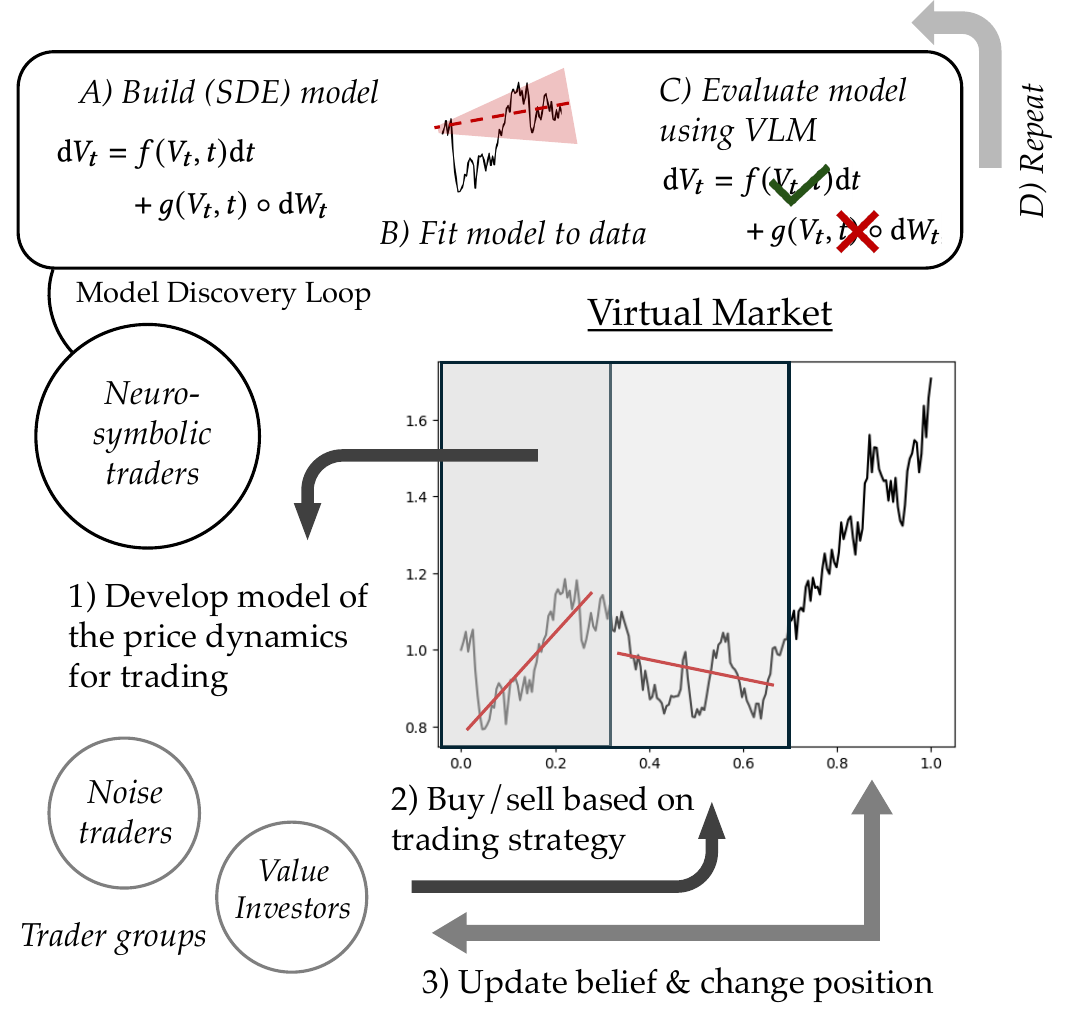}
\end{center}
\caption{Schematic for our neuro-symbolic traders that are exposed to a virtual market. A neuro-symbolic agent consists of a visual-language model (VLM) which can output model code for a stochastic differential equation (SDE) based on some observed price path image and instruction (A). The model is fit to the data using either a gradient-based or VLM-based calibration (B). Model output and observations are then passed to a second VLM which evaluates the model performance and suggests improvements (C). This is then passed to the original VLM and the loop continues (D). The steps within the virtual market are outlined in the figure.}
\label{fig:schematic}
\end{figure}

In this work, we develop neuro-symbolic agents, capable of adjusting their models of the world based on provided observations. In order to decouple the inference from model discovery, we follow \cite{li2024automated} in using differentiable programming to directly calibrate the models to data. However, initial parameter values and, importantly, changes to the model are driven by vision-language models (VLMs). Additionally, we use `Box's Loop' to formalise the model disovery process, with two agents, representing a critic, $C$, and model builder, $B$, that follow a chain of thought (CoT) prompting pattern to discover a stochastic differential equation that best matches the data \cite{wei2022chain}. We discuss the details for each of these components in the following sections and provide an overview of the components outlined in \autoref{fig:schematic}.

\subsection{Stochastic Models of the Fundamental Value}\label{sec:fund_val}

Our neuro-symbolic traders are tasked with determining the fundamental value that is believed to drive the price dynamics of a specific asset. We assume that the fundamental value, $V_t$, can be approximated by a stochastic differential equation (SDE), namely that
\begin{equation*}
\mathrm{d}V_t = f(V_t,t)\mathrm{d}t + g(V_t,t) \circ \mathrm{d}W_t, 
\end{equation*}
where $f(V_t,t)$ denotes the deterministic term, $g(V_t, t)$ denotes the stochastic term and we assume that $V_t$ is a continuous $\mathcal{R}^n$-valued process representing the evolution of price dynamics for the fundamental value of an asset. The stochastic term includes composition with Brownian motion, $\circ$ $\mathrm{d}W_t$, which we assume to be sampled from the Normal distribution with mean zero and standard deviation proportional to the time step, $\mathrm{d}W_t \sim \mathcal{N}(0, \mathrm{d}t)$ \cite{mao2007stochastic, revuz2013continuous}. 

The functional form of an SDE approximating financial time series is likely to be complex and time varying. This is because exogenous shocks to markets, such as interest rate changes or global crises such as the recent COVID-19 pandemic, are unlikely to be well-described when using historical data that does not contain these events. This is the ``black swan" problem of model selection in finance \cite{orlik2014understanding}. 

To account for this, we refer to classes of SDEs in general terms as $v_t \in \mathcal{V}$, where $\mathcal{V}$, denotes the super-set of SDE descriptions of the time series and $v_t$ is a concrete realisation, such as geometric Brownian motion, an Ornstein-Uhlbeck process, or even an SDE that partly or fully leverages artificial neural networks (NNs), $v_{t, \Theta}$ where $\Theta$ reflects the parameters of the NN \cite{kidger2021neural}. 

Throughout this work, we expand the set of possible functional forms, $\mathcal{V}$ to be large enough to include models with new independent variables and terms. This includes, for example, stochastic volatility (SV) models which describe the evolution of both the price dynamics as well as the volatility of the price. However, these additions are added by the LMs, using the process described in \autoref{sec:model_disc}. 

Here, we are interested in using language models (LMs) to discover the appropriate model of the fundamental value. To simplify implementation, we decouple model inference and calibration from model development. Hence, we use a differentiable model of the market dynamics, available through \texttt{diffrax} and which includes support for neural SDEs \cite{kidger2021on}. Model parameters, $\theta$
, are then calibrated directly to the historical data using gradient descent, given a pre-chosen set of summary statistics, $\mathbf{x}$,  and loss function, $\mathcal{L}(\mathbf{x})$. However, using gradient descent does not explicitly provide the posterior distribution over the calibrated model parameters as with simulation-based inference \cite{dyer2022black, cranmer2020frontier}. We leave this to future work. Additionally, using a differentiable simulator means that, where required, agents can approximate components of the market dynamics using NNs. This is also left to future work due to the complexity involved in effectively calibrating neural SDEs to the data.  

In this work, we first perform tests on a synthetic price signal generated by simple geometric Brownian motion (GBM), namely 
\begin{equation}
V_t = \mu x + \sigma W_t,
\end{equation}
where $\mu$ and $\sigma$ are the drift and volatility terms calibrated using gradient descent. This model is the starting model, $z_0$, for the discovery loop used by all neuro-symbolic traders and which we set out below. 

\subsection{VLM-based Model Discovery}\label{sec:model_disc}

The ability to determine whether a model is well-defined and suitable for describing some set of observations will depend on both epistemic and aleatoric considerations, namely whether the starting assumptions are correct and the degree of noise around the individual observations. However, the epistemic component of a model is extremely broad, covering both certainty around specific parameters, model structure and terms, and the existence of unaccounted latent or nuisance parameters. Recent work has shown that language models are able to perform automated model discovery \cite{li2024automated}. This early work highlights that VLMs can be used to identify models in both open and domain-restrictive search spaces as well as to improve existing models. 

Here, we consider the same problem of model improvement, focusing on determining the stochastic differential equations used to model financial time series. In order to use VLMs for this task, we follow \cite{li2024automated} in including both a critic and builder. Specifically, we use one VLM agent to evaluate the initial model fit and the second to build the model using the critic's evaluation. The coupled VLM-based program discovery is the combined neural (VLM) and symbolic (SDE) aspects of our neuro-symbolic traders. 

We provide the critic model, $C$, with a dataset, $\mathcal{D} = \{x_t,y_t\}_{t=0}^{T}$, consisting of $t$ observations, $y_t \in \mathrm{R}$ for input values $x_t$. This dataset is provided as a single image, normalised between 0 and 1 in both $x$ and $y$ values. We also assume there exists metadata, $\mathcal{C}\in \Sigma$, which provides additional context to the VLM and is a subset of the VLMs vocabulary, $\Sigma$. Here, metadata can vary from additional details on the starting model, such as whether the model has a jump diffusion term or details about the observations, such as what the financial time series represents. We then task the critic model with identifying improvements to the model, 
\begin{equation}
c_i \sim q_{\mathrm{LM},C}(\cdot | z_0, z_1, \dots, z_{i-1}, h_C, \mathcal{D}).
\end{equation}
where $c_i \in \Sigma^C$ is the VLM-based evaluation of the model, drawn from the vocabulary of the critic model, $z \in \mathrm{Z}$ is the model description, where we pass all previous model descriptions in each pass and where $z_0$ is GBM as discussed in \autoref{sec:fund_val}, $h_C \in \Sigma^*$ is the natural language instruction for the critic VLM that we use to combine previous modelling approaches and guide the VLM, and $\mathcal{D}$ is the latest description of model success which is an image with historical observation overlaid by the output from the SDE.

We then pass this evaluation to the builder agent, $B$, to suggest new changes to the model, namely
\begin{equation}
z_i \sim q_{\mathrm{LM},B}(\cdot | c_i, h_B),
\end{equation}
where  $h_B \in \Sigma^*$ is the natural language instruction for the builder. Here, we provide the dataset and model history only to the critic to simplify the overall process. The newly suggested model, $z_i$, is then calibrated to the data using gradient descent, as described in \autoref{sec:fund_val}. 


We test three different implementations of our neuro-symbolic agents, with increasing complexity. We first ask the critic to provide only starting parameter values, $\theta_0$ using model terms, $z_i$, and dataset, $\mathcal{D}_i$. In a second test, we ask the agents to suggest model improvements, $z_{i+1}$. Finally, we provide additional domain-specific details about the data, $\mathcal{\bar{D}}$, including the source of the financial time series and time period. 

After conducting these tests, we expose groups of these agents to a virtual market where they make trading decisions based on their belief of the evolution of the price path. For all results, we use the \texttt{Claude-3.5-Sonnet} model from Anthropic \cite{Anthropic2024Claude35Sonnet}. All results in this work are with the temperature of the VLM set to 0.  

\begin{figure*}[!t]
    \begin{subfigure}{0.24\textwidth}
        \includegraphics[width=4cm]{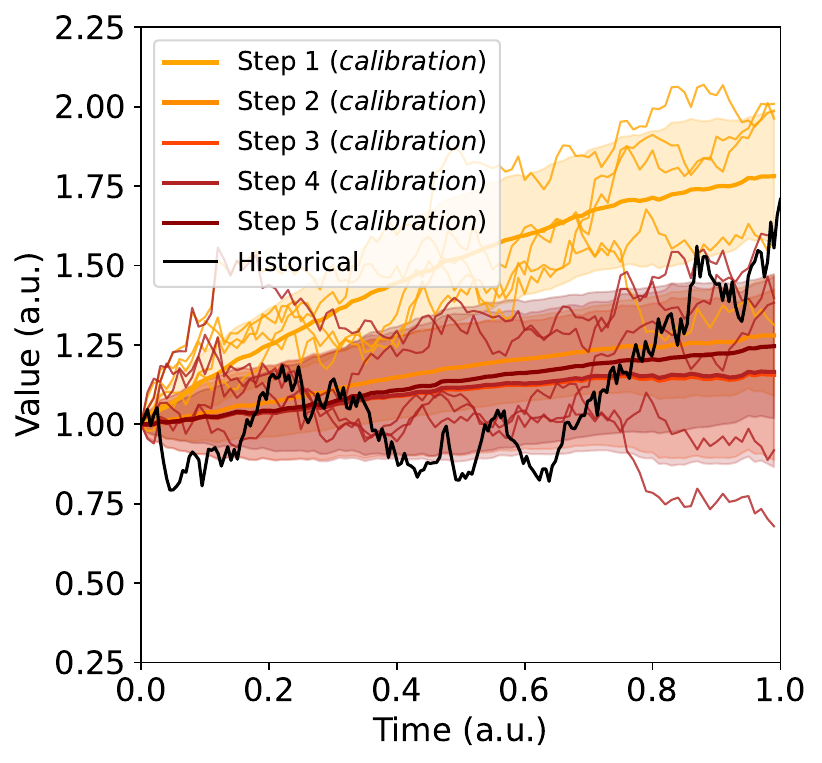}
        \caption{VLM calibration}
    \end{subfigure}
    \begin{subfigure}{0.24\textwidth}
        \includegraphics[width=4cm]{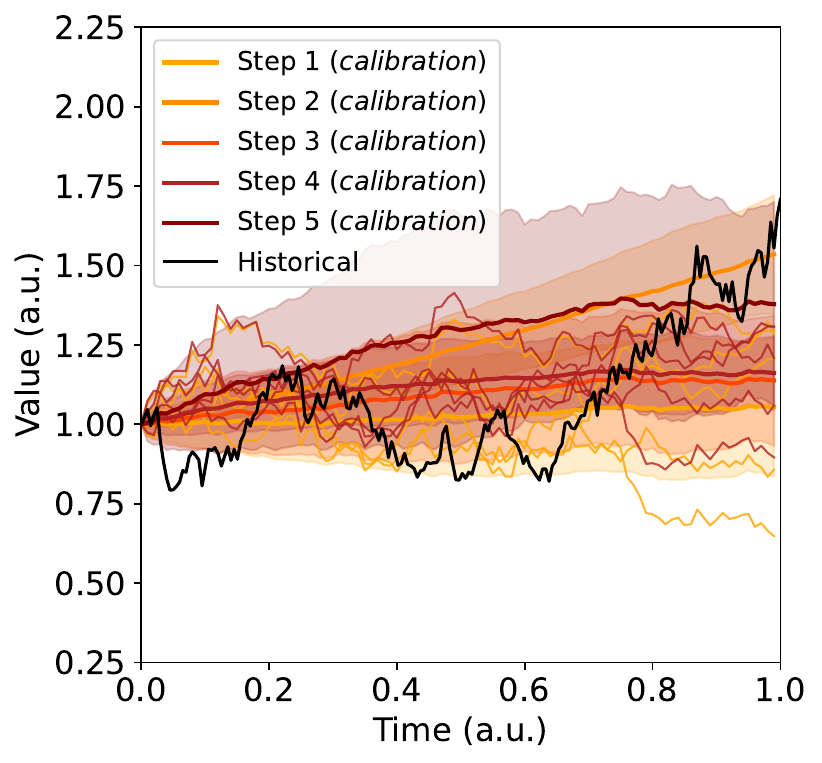}
        \caption{Gradient calibration}
    \end{subfigure}
    \begin{subfigure}{0.24\textwidth}
        \includegraphics[width=4cm]{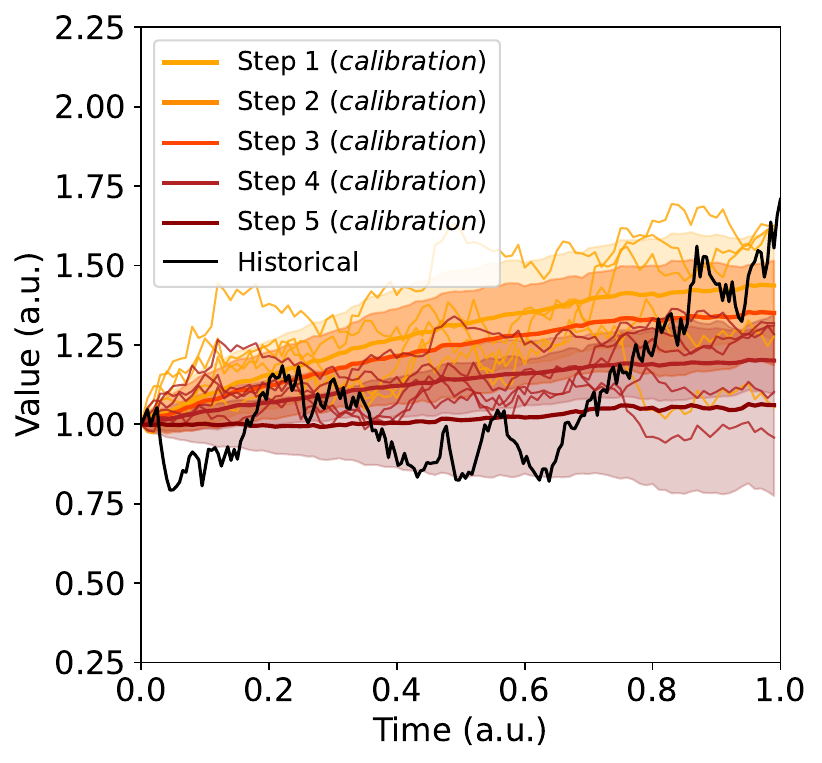}
        \caption{VLM calibration \& `brevity'}
    \end{subfigure}
    \begin{subfigure}{0.24\textwidth}
        \includegraphics[width=4cm]{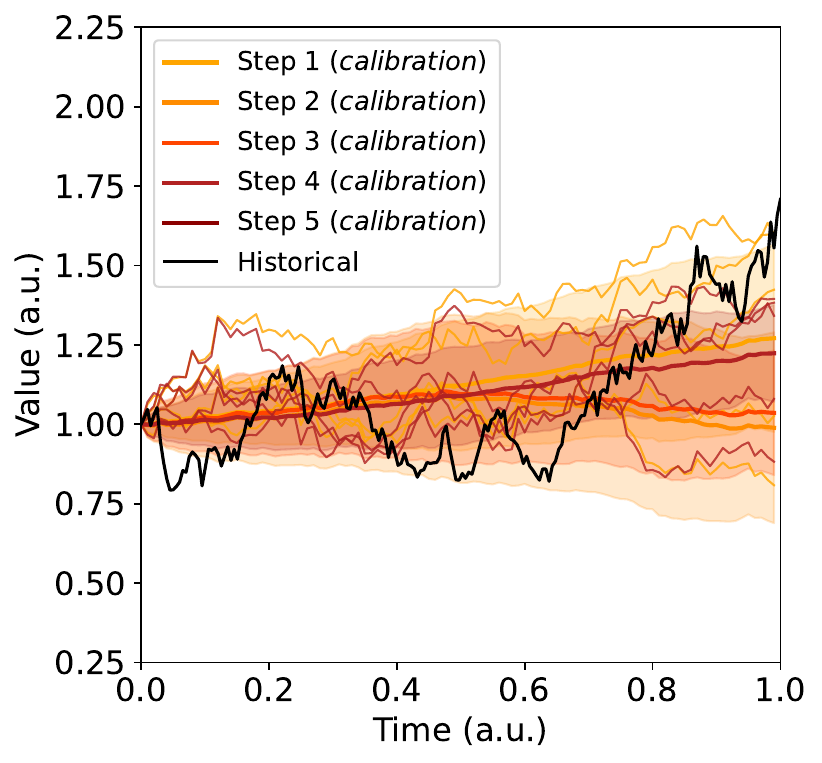}
        \caption{Gradient calibration \& `brevity'}
    \end{subfigure}
    \caption{Comparison of different methods used to fit a stochastic differential equation to time series data, including (a) a vision-language model (VLM) approach where parameter values are suggested based on the image and model description, (b) a gradient-based approach where the error is directly minimised using differentiable programming, and (c-d) which introduces additional instructions to the VLM to construct simple or parsimonious models.}
    \label{fig:gbm_disc}
\end{figure*}

\subsection{Market Simulation}\label{sec:market_sim}

Given the model of fundamental value, described in \autoref{sec:fund_val}, and the model discovery process, described in \autoref{sec:model_disc}, we next set out how an agents choice in fundamental value model can influence the price dynamics. Given that the belief of the fundamental value of a stock is typically coupled to the price of a stock, we assume that the fundamental value can be at least partly approximated by the price. However, in practice, the price dynamics of a specific asset will be driven by the shared interaction of different traders that are expected to have different beliefs on the fundamental value. The interplay of these traders occurs at the market-level such that the market mechanism performs an aggregation operation across the individual traders beliefs. 
We are interested in the effect on the price dynamics for multiple agents that estimate the fundamental value using neuro-symbolic methods. Specifically, we are interested in the feedback in model decisions made by the neuro-symbolic agents, given changes to the price dynamics caused by other agents in the market using the same methods. In order to quantify these effects, we construct a virtual market where we can simulate the price dynamics according to the demand from groups of different traders. In this work, we use a simplified version of that set out in \cite{majewski2020co}, consisting of two trader types, fundamental traders and noise traders. 

The fundamental or value investors are described by, 
\begin{equation*}
Q^F_t = \kappa (V_t - P_t) 
\end{equation*}
where the change in demand for fundamental investors, $Q^F_t$, is driven by the mismatch between the price $P_t$ and the perceived fundamental value $V_t$, according to some sensitivity $\kappa$. 

Here, we assume that individual demand, $Q^{i,F} \in Q^F$, is representative of a group of traders that are making decisions based on a similar shared belief rather than a single individual trader. 

We also assume that there is a large group of other traders who make decisions which cannot be obviously attributed to changes in the fundamental value, which we refer to as noise traders $N_t$. For the noise trader, this is defined by
\begin{equation*}
Q^N_t =\sigma dW_t,
\end{equation*}
where $dW_t$ is a Wiener process and where $\sigma$ describes the overall noise level introduced by noise traders, $Q^{i,N} \in Q^N$. The total demand in the market is assumed to be the aggregation of these two different trading classes, namely,

\begin{equation}\label{eq:demand}
Q_t = \overbrace{\kappa (V_t - P_t)}^{\text{fundamental traders}} + \overbrace{\sigma dW_t}^{\text{noise traders}}. 
\end{equation}

\begin{figure*}[!h]
    \begin{subfigure}{0.3\textwidth}
        \includegraphics[width=5cm]{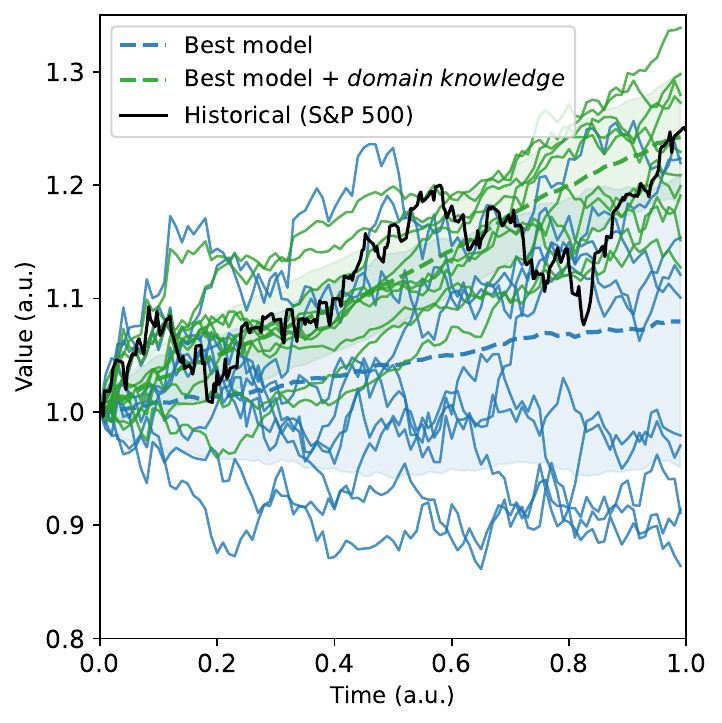}
        \caption{S\&P500}
    \end{subfigure}
    \begin{subfigure}{0.3\textwidth}
        \includegraphics[width=5cm]{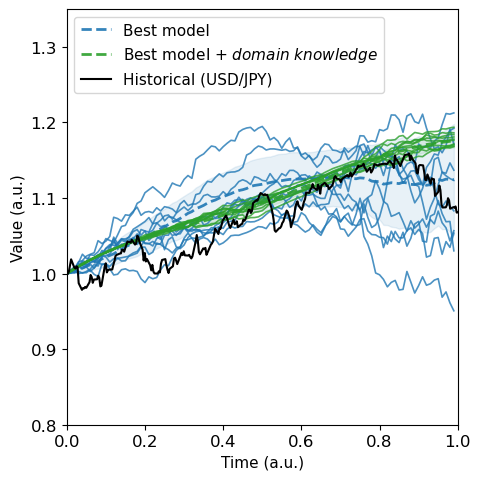}
        \caption{USD/JPY}
    \end{subfigure}
    \begin{subfigure}{0.3\textwidth}
        \includegraphics[width=5cm]{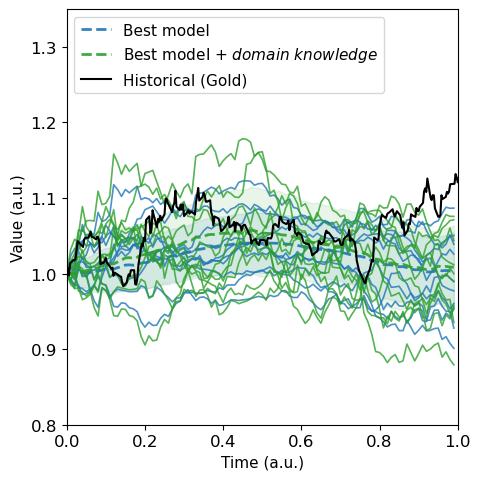}
        \caption{Gold}
    \end{subfigure}
    \caption{Comparison of models for the fundamental value proposed by a neuro-symbolic trader for three different asset classes and comparing both with and without domain information passed to the trader.}
    \label{fig:asset_models}
\end{figure*}
Hence, the overall price dynamics evolves according to the aggregated demand and this, in turn, is determined by the mismatch between the current price and the expected price based on beliefs around the fundamental value as well as some noise. We assume a linear price impact model, also known as Kyle's lambda $\lambda_K$, such that the evolution of the price dynamics is given by
\begin{equation}\label{eq:price}
dP_t = \lambda_K Q_t.
\end{equation}


For all results, we assume the demand from noise traders is sampled from a normal distribution with scale 0 and standard deviation 10$^{-1}$. We also fix Kyle's lambda as 10$^{-1}$. In our results, we test three fundamental trader groups and a noise trader group. 

\section{Results}\label{sec:results}
In what follows, we describe three experiments that demonstrates both the functionality of our neuro-symbolic agents and the impact of exposing these to a market place. First, we demonstrate that the neuro-symbolic agents are able to identify fundamental value models using synthetic data. We then test the neuro-symbolic agents on real world data. Finally, having validated our neuro-symbolic agents using both synthetic and real world data, we assess the impact of financial markets that contain neuro-symbolic traders.

\subsection{Identifying Fundamental Value Models}\label{sec:synth}
The ability of neuro-symbolic traders to identify a potential fundamental value signal is key to our virtual markets. Here, we note that the purpose is to evaluate the types of models and the overall complexity of models discovered by the neuro-symbolic traders. We use a ground-truth model of geometric Brownian motion (GBM) and normalise both time and price values to be between zero and one to minimise any unintended information parsed to the agent and improve stability of the model discovery process. For all tests, we fix values of the number of simulations to run as 100, the starting location for the signal (which is 1), and the time resolution $\mathrm{d}t= 0.01$.  

\subsubsection{Parameter estimation}

We first investigate how the model performs at parameter estimation. Here, we consider two different methods. As the stochastic equations are executed using the \texttt{diffrax} package \cite{kidger2021on}, which extends the differentiable programming \texttt{JAX} library \cite{jax2018github}, we can use gradient descent to directly calibrate model parameters from the data. For all experiments, we use a learning rate scheduler with exponential decay and decay rate of 0.9. To account for stochasticity within the model, we use the first four moments of the data as summarising statistics, weighting the first two moments higher to steer the model towards matching the mean and volatility of the price. We run the calibration routine for 100 epochs, with gradient clipping (with threshold of 5) and an L2 regularization penalty added to the loss with strength 10$^{-5}$ which we find improves stability of the calibration process. We find that this calibration procedure is able to identify the parameters which reproduce these moments with an mean absolute error (MAE) of $(7.30 \pm 0.43) \times 10^{-2}$. We note that we focus on the accuracy of the moments for all our evaluations to remain consistent with the model discovery loop where the number of parameters is expected to vary.

As discussed in \autoref{sec:model_disc}, we use a vision language model (VLM) as the critic for model evaluation. This allows us to also directly calibrate the model using the VLM. Here, we pass the VLM component of the agent a figure that highlights the historical data (black) and the output of the model overlaid (represented as coloured lines). We also provide the model code. We then prompt the VLM to suggest relevant parameters for improving the model accuracy. The ability for VLM and other multi-modal models to act as a calibration engine is an open question. Here, we find that the model is ability to identify the parameters which reproduce these moments with an MAE of $(8.42 \pm 1.04) \times 10^{-2}$. This performance is lower but significantly faster than the previously described calibration process. However, due to the inherent stochasticity of the VLM, highlighted by the larger standard deviation, we use the gradient-based calibration process in what follows. 

\subsubsection{Suggesting model terms}

One of the main aims of this work is assessing the quality and diversity of models discovered by the neuro-symbolic traders. Specifically, we are interested in quantifying the extent to which these models converge to a common model formalism. To assess this, we ask the neuro-symbolic trader to evaluate model terms, suggest new ones, and then repeat this critique-build-discover loop for a number of steps. We perform the loop three times, with five rounds of model discovery.

We initialise all models with the same starting model (GBM) to ensure that the format and structure of the code is correct. This means that we are, in effect, providing the neuro-symbolic agent with the correct model from the start. However, across all experiments and iterations of the model discovery loop, we find that the neuro-symbolic agent consistently increases the complexity of the model. In total, there are ten broad groups of different model extensions. The most commonly introduced are a square root drift term (mentioned 11 times), mean-reversion term (mentioned 9 times), sinusoidal term (5 times) and time-dependent volatility scaling (3 times). Some of the less frequently suggested model improvements include an adaptive drift term (mentioned once) and logistic growth term (mentioned twice). There were also 5 variations on additional non-linear volatility terms. 

In \autoref{fig:gbm_disc}, we show example output of the model over five rounds of model discovery. As can be seen, the overall evolution of the model is varied but the final model result does closely resemble the original time series. Furthermore, the error in the calculated moments decreases from $7.7 \times 10^{-2}$ to $7.0 \times 10^{-2}$ when using gradient-based calibration or from $8.4 \times 10^{-2}$ to $8.2 \times 10^{-2}$ when using the VLM directly. This indicates that the neuro-symbolic agent has effectively identified a plausible model for the underlying asset.

For all iterations, we find that the model gets consistently more complex with each step of the model discovery loop. Specifically, we note that one or two terms are added in each step, resulting in models that have 7-8 parameters. In order to reduce the complexity and steer the VLM towards simpler expressions within the model space, we also tested passing the VLM different instructions, $h_\mathcal{C}$ and $h_\mathcal{B}$, which explicitly requested brevity in the model code and to build  parsimonious models. For these iterations, we again test both with and without gradient-based calibration. 

We find that the model terms do reduce in complexity when adapting the instructions. The number of different groups of terms is now 8 and the total number of parameters in the final model is between 5 to 6. The models continue to contain a mean-reverting component (10 times), including one instance of a long-term mean reversion term, as well as a square root diffusion term (8 times). However, we find a large number of models contain just a drift term with a constant (9 times). There is now only one instance of a sinusodial or periodic term added to the model. 

Despite, or potentially because of, the simpler model description, we find that the error increases. Interestingly, we observe that calibrating the model using the VLM directly decreases the MAE to $(7.94 \pm 0.53) \times 10^{-2}$ compared to $(9.41 \pm 4.56) \times 10^{-2}$. This is possibly due to the added consideration given to the model development by the VLM, similar to the ``think step-by-step" prompt pattern. Overall, these results highlights the sensitivity of VLMs to their instructions, $h_i$. We show the final result identified by the `parsimonious' model-discovery loop in \autoref{fig:gbm_disc}. 
 


\subsection{Testing on Financial Time Series}\label{sec:assets}

In the previous section, we assessed how neuro-symbolic agents might discover a model of time series data. Next, we consider how these agents perform specifically within the context of financial time series and whether there is a difference in performance across different asset classes. To test this, we consider the historical period between the first of January 2023 to 2024 for three different asset classes; an equity (S\&P 500), commodity (gold), and foreign currency exchange (USD$\backslash$JPY). We describe our results below. 

\subsubsection{Different Asset Classes}

For each asset class, we perform the model discovery loop five times, with five iterations for each loop. For all assets, we find that the neuro-symbolic agent is typically able to identify a meaningful model of the fundamental value. However, we note that the VLM will, on occasion, hallucinate terms into the model description which will cause the model to fail. This occurred 6 times over the 75 different iterations. 

For all successful iterations, we use the gradient-based calibration process and find the overall error across all assets to be $(3.11 \pm 0.4) \times 10^{-2}$. Additionally, we see a small decrease in MAE for the last iteration compared to the first, where the MAE for the first iteration is $(3.26 \pm 0.02) \times 10^{-2}$. Overall, we find the accuracy of the model increases with each iteration of the model discovery loop. 

In \autoref{fig:asset_models}, we show the models discovered for the three different asset classes. For each asset class, we find that the model complexity increases with each iteration, as in \autoref{sec:fund_val}. Overall, we find that the number of parameters for the first iteration is typically three, where the model adds a  mean-reversion rate to the volatility and drift coefficients. However, by the fifth iteration, the model complexity has grown to 6 or 7 parameters.

Comparing across asset classes, we observe that the same model terms and parameters are typically introduced and these do not significantly differ from those highlighted in \autoref{sec:fund_val}. Overall, we see that models tend to add in mean-reversion rate (12 times), long-term mean (10 times) and a seasonal component (12 times). 

Despite this, we observe difference for the individual asset classes. We find that the neuro-symbol traders are able to identify models with better accuracy (lower MAE) for gold ($2.8 \times 10^{-2}$), but considerably worse for both the S\&P 500 ($3.4 \times 10^{-2}$) and USD/JPY ($3.5 \times 10^{-2}$. Given the model has no other information other than the price signal, this may be because the gold price series is flatter than the other assets over this period. We note that the neuro-symbolic traders consistently introduce a non-linear growth term for the USD/JPY time series, and rarely for the other assets. We also observe that there is a consistent introduction of a seasonal term, possibly related to the periodicity observed in the time series. 

These results indicate that the price signal itself is sufficient to introduce some, small, variation in the model discovery process. This highlights that our neuro-symbolic traders, at least \textit{prima facie}, adjust their model discovery process to the observations. 

\begin{figure*}[!t]         
\includegraphics[width=18cm]{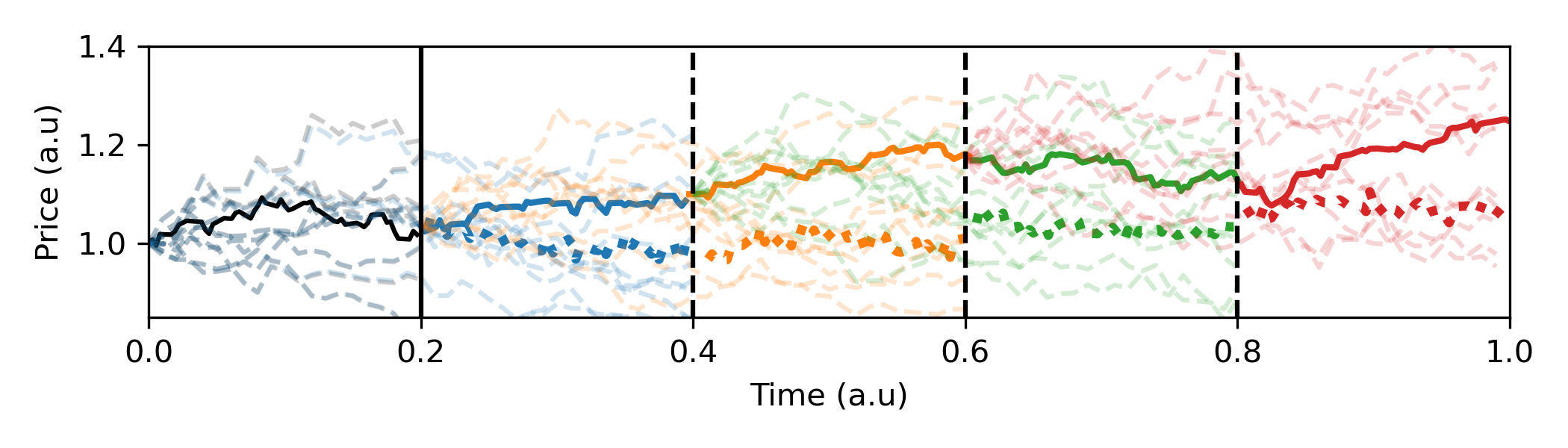}
    \caption{Simulated price path (dotted line) generated by neuro-symbolic traders model of the fundamental value (faded lines) within a virtual market. Each section, marked by a different colour and separated by vertical lines, relates to periods used to fit the model ($t_{i-1}$) and then simulate the impact in price ($t_i$). The original S\&P 500 stock is included for reference (solid line). }
    \label{fig:market_sim}
\end{figure*}

\subsubsection{Adding Domain Knowledge}

To further investigate whether these neuro-symbolic traders introduce asset-specific model terms, we next consider the influence of domain knowledge in the model discovery process. 

We now use the same process of model discovery outlined in \autoref{sec:fund_val} and \autoref{sec:model_disc}, but make an additional change to the VLM instructions to account for the specifics of the financial time series data by amending the metadata, $\mathcal{C}$, to include both the asset and trading period and explicitly state that this is financial time series data. In \autoref{fig:asset_models}, we show the models discovered after adding domain knowledge to the neuro-symbolic agent. 

We find that adding in domain knowledge to the neuro-symbolic trader alters the model discovery process, introducing more robustness in the discovered models (no fails in model execution) and more diversity in the final model description, with 6 to 10 params in the final model. We also find that the accuracy of the model improves with a lower MAE in both the first model discovery step, $(2.75 \pm 0.04) \times 10^{-2}$, as well as in the last, $(2.67 \pm 0.06) \times 10^{-2}$. As with previous results, we see the overall MAE decrease with each iteration of the model discovery loop. 

Considering each asset in turn, we again observe differences in both model accuracy and complexity. In terms of accuracy, we now observe that the neuro-symbolic trader identifies all assets with similar MAE. The USD/JPY exchange rate has slightly lower MAE, $(2.47 \pm 0.1) \times 10^{-2}$, whereas the S\&P 500 and gold price data are $(2.81 \pm 0.6) \times 10^{-2}$ and $(2.72 \pm 0.06) \times 10^{-2}$, respectively. We note that both with and without domain knowledge results in similar models identified for the gold price signal.

When considering the individual assets, we observe more variation in model terms. This includes an exogenous temporary shock term for the S\&P 500 and a jump diffusion term for both USD/JPY and gold. Overall, we see model convergence for the first 2-3 iterations of the model discovery loop, with almost all identified models containing mean reversion and a long-term mean term. For both the S\&P 500 and gold, there are seasonal components. For almost all iterations specific to the S\&P 500, there is also some nonlinear or logarithmic growth component. 


\subsection{Neuro-Symbolic Traders in Markets}\label{sec:hist}

We have demonstrated that our framework for neuro-symbolic trader agents are able to identify stochastic models for financial time series using only the price data. 

However, all results thus far treat the agent as an isolated instance. Given the increasing popularity of deep generative models, it is possible that methods using a broadly similar approach to price discovery becomes increasingly applied to trading strategies. In this section, we consider the influence of groups of neuro-symbolic traders (the `AI crowds') on financial markets. 

To quantify the influence of these groups on financial markets, we follow the description given in \autoref{sec:market_sim} and introduce three neuro-symbolic traders into a virtual market. We use the same S\&P 500 data as used in \autoref{sec:model_disc} but split into five independent sections covering $t_{i-1}$ to $t_i$ for $i = \{1, 2, 3, 4, 5\}$. We provide the traders with domain-specific information on both the asset class and trading period. We use ten realisations of the fundamental value model from each trader, such that each trader can be considered as acting as a trader group, and where using too many realisations may average out the stochasticity within the individual models. 

To simulate the price path, we task the neuro-symbolic traders with estimating a model for the fundamental value of the asset for each $(i-1)^\mathrm{th}$ section and use the estimated fundamental price as the signal, $V_t$, passed to the fundamental value agents in \autoref{eq:demand}. This allows us to calculate the price signal, \autoref{eq:price}. We then pass this updated price signal to the separate traders to estimate the fundamental value for the $(i-1)^\mathrm{th}$ section and continue until we have coverage over the total period. This process can be understood as a form of moving window simulation, where the model discovery process is used to estimate the fundamental value which, in turn, informs the price dynamics for the subsequent section. Results from this experiment are shown in \autoref{fig:market_sim}.

We run five trials, exposing the three neuro-symbolic traders to a virtual market. Across all trials, we find that the price signal significantly flattens, as shown in \autoref{fig:market_sim}, when compared to the original S\&P 500 price. This is despite an increasing complexity of the underlying fundamental value, which follows the same rise in complexity as outlined in both \autoref{sec:synth} and \autoref{sec:assets}, with an increase in both number of terms and corresponding parameters.

We find that the number of parameters increases from 3 to 6-7 at the end of the simulation. Additionally, we observe that the traders are far more likely to make more complex changes to the model such as introducing jump diffusion terms or stochastic volatility. Additionally, we observe that the MAE fluctuates between $3 \times 10^{-2}$ to $5 \times 10^{-2}$, with no clear trend across multiple iterations or trials. 

These results indicate that markets containing neuro-symbolic traders may result in less reactive price signals and crowding of price movements, despite diversity in the underlying model of the fundamental value. 

\section{Discussion \& Future Work}\label{sec:discuss}

We have set out how the combination of deep generative models and symbolic models, specifically stochastic differential equations, can be used to determine a simple trading strategy based on their belief on the fundamental value of an asset, which we describe as neuro-symbolic traders. By combining these traders with a model-discovery loop, we show that they are able to identify reasonable models for the price dynamics both when applied to synthetic data,  described in \autoref{sec:synth}, and historical observations of different financial assets, described in \autoref{sec:assets}. 

We make two important observations from these results. First, we see that increasing iterations typically improves the overall model performance, with a decreasing mean absolute error (MAE). Second, we find that these models increase in complexity, evidenced by an increase in model parameters and terms. These results support our claim that these neuro-symbolic traders are effective at discovering and improving upon models of the fundamental value of an asset.

Given that a key component of our neuro-symbolic traders is the use of a vision-language model (VLM), we also explore the influence of different instructions passed to the VLM. In \autoref{sec:synth}, we alter the instructions to the VLM to guide it towards parsimonious and brief model descriptions. We find that the discovered model reduces in complexity, as expected, but that this has a corresponding increase in the MAE. Then, in \autoref{sec:assets}, we add in domain-specific details. Here, we find that the discovered models both increase in complexity, with a wider range of different terms tried, and that this results in a lower MAE. These results highlight both the sensitivity of the instructions to neuro-symbolic traders but also their adaptability to contextual data. 

Our results demonstrate the potential value in applying neuro-symbolic methods to the discovery process of financial models. We identify two promising aspects where we can extend this work. First, we believe that further information could be parsed to the neuro-symbolic agents, such as technical analysis on the time series or additional asset-specific data. This would increase the domain specificity of the model and, we believe, have corresponding increase in performance. Second, we have not tested the use of neural components within the stochastic model of the fundamental value. This would likely further increase the accuracy of the model, but with the trade-off that there is less interpretability in the choice of model terms. 

Having demonstrated our framework for neuro-symbolic traders, we expose them to a virtual market environment. Surprisingly, we find that this causes the price signal to flatten, with lower volatility and less price movement. This indicates that exposing neuro-symbolic traders into market environments may lead to herding of prices, despite diversity across the model of the underlying asset. Future work will consider the effect of heterogeneous neuro-symbolic traders which act based on different information. Additionally, it would be interesting to consider how these traders react to sudden shocks in the price, approximating `black swan' events.  

We believe our work gives important early empirical evidence on the impact of exposing deep generative models into markets. We have outlined a simple framework for building and testing neuro-symbolic traders. However, the main impact will be the continued improvement in reasoning capabilities of AI models which continues to have ramifications across the fields.

\section{Significance}

Our work sets out a framework for developing neuro-symbolic trader, which combine multi-modal deep generative models with stochastic modelling of financial time series. We demonstrate our traders on both synthetic and historical time series and find good performance. We then expose our traders into a virtual market environment and find that they cause price suppression, despite diversity in their belief in the fundamental value of the underlying. We believe that our work gives critical information on both the construction and implications for neuro-symbolic methods in financial markets, with impact to regulatory frameworks, trading strategies, and market stability. 



\bibliographystyle{ACM-Reference-Format}
\bibliography{main}

\end{document}